\documentclass[letterpaper, 10 pt, conference]{ieeeconf} 
\IEEEoverridecommandlockouts
\makeatletter
\let\NAT@parse\undefined
\makeatother
\usepackage[numbers,sort&compress]{natbib}
\usepackage{amsmath,amssymb,tabularx,graphicx,flushend}

\title{\LARGE \bf
Fast Deep Stereo with 2D Convolutional Processing of Cost Signatures%
}

\author{Kyle Yee$^{1}$ and Ayan Chakrabarti$^{2}$
\thanks{$^{1}$Kyle Yee is an undergraduate student at Swarthmore College, Swarthmore, PA 19081, USA
        {\tt\small kyle.g.yee@gmail.com}}%
\thanks{$^{2}$Ayan Chakrabarti is with the Department of Computer Science \& Engineering, Washington University in St. Louis,
        St. Louis, MO 63130, USA
        {\tt\small ayan@wustl.edu}}%
    }
    
\usepackage[colorlinks=True]{hyperref}
\makeatletter
\hypersetup{pdftitle={\@title},pdfauthor={Kyle Yee and Ayan Chakrabarti}}
\makeatother

\begin{document}

\maketitle\thispagestyle{empty}\pagestyle{empty}

\begin{abstract}
  Modern neural network-based algorithms are able to produce highly accurate depth estimates from stereo image pairs, nearly matching the reliability of measurements from more expensive depth sensors. However, this accuracy comes with a higher computational cost since these methods use network architectures designed to compute and process matching scores across all candidate matches at all locations, with floating point computations repeated across a match volume with dimensions corresponding to both space and disparity. This leads to longer running times to process each image pair, making them impractical for real-time use in robots and autonomous vehicles. We propose a new stereo algorithm that employs a significantly more efficient network architecture. Our method builds an initial match cost volume using traditional matching costs that are fast to compute, and trains a network to estimate disparity from this volume. Crucially, our network only employs per-pixel and two-dimensional convolution operations: to summarize the match information at each location as a low-dimensional feature vector, and to spatially process these ``cost-signature'' features to produce a dense disparity map. Experimental results on the KITTI benchmark show that our method delivers competitive accuracy at significantly higher speeds---running at 48 frames per second on a modern GPU.
\end{abstract}

\section{Introduction}
\label{sec:intro}

The availability of real and synthetic datasets~\cite{kitti,kitti2,sceneflow} and use of deep neural networks~\cite{zbontarlecun2,iresnet,pds,pyramid} has made stereo estimation increasingly reliable. As evidenced by their performance on realistic benchmarks~\cite{kitti2}, modern algorithms are able to produce depth estimates from stereo image pairs with reliability that nearly matches depth measurements from more expensive devices such as LIDARs. However, a significant roadblock to practically using these algorithms for depth perception in robots and autonomous vehicles is their computational expense. While traditional stereo methods were able to generate dense depth estimates in real time, albeit with lower accuracy, modern neural network-based stereo methods take more than half a second (often much more) to process a single stereo pair at a standard resolution.

The processing pipeline of a stereo algorithm has two computational components: computing a matching cost volume based on similarities between all pairs of reference and matching candidate points in the stereo pair, and processing this volume to yield robust depth estimates by reasoning about smoothness, planarity, etc. in natural scenes. Since the cost volume itself is large (of size equal to number of pixels times the number of candidate disparity values), traditional stereo algorithms emphasized efficient operations for both constructing~\cite{census} and processing~\cite{sgm} this volume.

\begin{figure}[!t]
  \centering
  \includegraphics[width=\columnwidth]{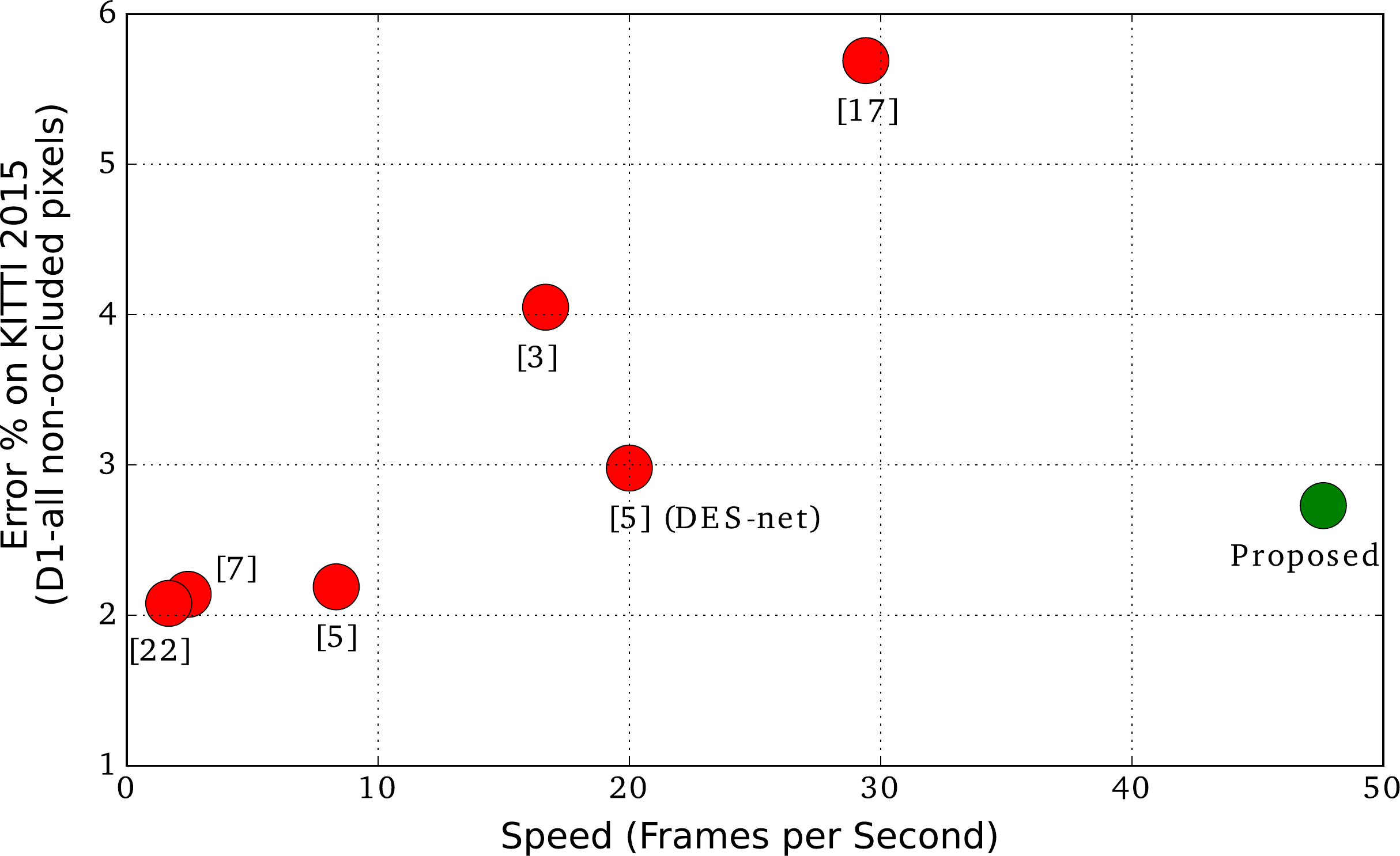}
  \caption{Accuracy-Speed Trade-off in Stereo Estimation. We show the error rate vs speed of different stereo algorithms on the KITTI 2015~\cite{kitti2} benchmark. Our method yields accuracy competitive with state-of-the-art networks for stereo estimation while being significantly faster, and therefore practical to deploy on real systems.}
  \label{fig:teaser}
\end{figure}

However, neural network-based methods must instantiate these computations using cascades of layers and hence incur significant expense due to the large number of floating point operations repeated across the three-dimensional cost volume. Recent methods cast these as three-dimensional (3D) convolutions~\cite{iresnet,pds,pyramid} to improve parallelism and data flow, but they are still much slower than traditional stereo methods, and indeed, than typical neural networks that use only two-dimensional (2D) convolutional to process regular images.

In this paper, we propose a new neural network-based method for accurate stereo estimation with an architecture highly optimized for computational efficiency. Our approach is motivated by the success of methods for depth completion~\cite{s2d} that are able to reconstruct highly accurate depth maps given a small number of sparse depth measurements and a reference color image as guide. This demonstrates that natural scenes have significant structure that deep neural networks can learn to exploit, despite being given only weak or noisy geometric depth cues, suggesting that expensive computations for precise matching may be un-necessary.

Our method first constructs an initial cost volume using multiple traditional matching costs that are efficient to compute. It then converts the set of all scores for different disparities at each pixel to a low-dimensional ``cost-signature'' feature vector. This conversion is learned as a set of independent per-pixel layers that produce a succint summary of the stereo depth information at each pixel location. An encoder-decoder network then uses 2D convolutions to process this 2D feature map, rather than the more expensive 3D convolutions on a cost volume, and produces an estimate of the final disparity map. The conversion and encoder-decoder layers are learned jointly with end-to-end training.

Experimental results on the standard KITTI~\cite{kitti2} benchmark demonstrate that our models produces estimates with accuracy that is only slightly worse than the state-of-the-art, and higher than traditional stereo methods. But crucially, this accuracy comes at very low computational cost: our network is able to process stereo pairs at \emph{48 frames per second} on a modern GPU. As shown in Fig.~\ref{fig:teaser}, our approach affords a favorable trade-off between accuracy and speed, making it both reliable and practical for deployment in robots.

\section{Background and Related Work}
\label{sec:rw}

Depth estimation using stereo images from a calibrated camera pair requires establishing correspondences between pixels in the two images by searching over epipolar lines. When the cameras are related by only horizontal translation (or the images have been rectified to simulate this setup), all epipolar lines are horizontal and the problem reduces to finding the horizontal shift, or \emph{disparity}, between the $x-$ co-ordinates of the projected location of a surface point in the image pair. Stereo estimation is typically cast as the problem of estimating a dense disparity map---the value of disparity at every location in the co-ordinate system of one of the two images chosen as reference. Scene depth can then be derived from disparity given knowledge of relative camera poses.

Estimating disparity by finding dense correspondences is challenging due to the presence of smooth regions, repeating textures, specular highlights, and half-occlusions. Stereo algorithms proceed by first computing an initial score of match quality between each pixel in the reference pixels and all candidate matches. These candidates are indexed by a finite discrete set of candidate disparity values common to all pixels---typically integer pixel disparities ranging from zero to some maximum value---and correspondingly, the matching scores are organized in a cost ``volume'' along the spatial and disparity dimensions. Traditionally, stereo algorithms used hand-crafted similarity metrics for matching that take into account local neighborhoods around pixels for robustness, while also being efficient to compute~\cite{census}.

However, these matching scores are still ambiguous and thus local reasoning alone is insufficient for accurate disparity estimation. This is why stereo algorithms have a second ``globalization'' stage, where the local match information in the cost-volumes is aggregated while promoting properties such as smoothness, piece-wise planarity, etc.\ in the estimated disparity maps. This aggregation was traditionally as optimization of an energy function~\cite{sgm,SPS-St,chakrabarti2015low}, again with an emphasis on computational efficiency.

\u{Z}bontar and LeCun~\cite{zbontarlecun1,zbontarlecun2} demonstrated that using deep neural networks for stereo estimation could deliver significant improvements in accuracy over traditional stereo pipelines. Their work only replaced the local matching stage---they proposed learning networks that took a pair of $9 \times 9$ patches in the left and right image as input to produce a matching score. Once this network was trained, it was applied on all candidate match pairs to populate the cost volume, which was then smoothed using traditional aggregation techniques~\cite{sgm}. Surprisingly, by just replacing the matching cost with a learned metric, this method was able to achieve significant gains in accuracy. However, these gains came with a reduction in speed, taking more than a minute to process a single stereo pair. \u{Z}bontar and LeCun~\cite{zbontarlecun2} also considered faster architectures, as did Luo \emph{et al.}~\cite{siamese}, but these were less accurate and still took 0.7~\cite{siamese} and 0.8~\cite{zbontarlecun2} seconds to process a stereo pair.

These methods were driven by the presence of moderate-sized real stereo datasets~\cite{kitti,kitti2} with ground-truth data captured using a LIDAR. Noting the benefits of learned methods for stereo, Mayer \emph{et al.}~\cite{sceneflow} introduced a much larger, synthetically rendered, dataset to enable training of more complex networks---with layers that carry out both matching and globalization computations (the latter replacing traditional aggregation) and are trained end-to-end. GC-Net~\cite{3dconv} uses shared 2D convolutions to extract features from each image in the stereo pair, concatenates them to form a 4D tensor (indexed by spatial dimensions, disparity, and features) and then uses 3D convolution layers to process this cost volume. Since then, a number of methods have used a similar approach to using 3D convolutions with innovations in network architecture for cost-volume construction and processing~\cite{edgestereo,pyramid,pds,iresnet}, yielding improvements in accuracy and run-times. The fastest among these is the smaller DES-net architecture in \cite{iresnet}, which primarily allocates layers for accurate cost computation and achieves a run-time of 0.05 seconds per stereo pair.

The goal of our work is accurate but real-time stereo estimation, which we achieve through the use of traditional matching costs and 2D (instead of 3D) convolution layers. In this context, it is useful to discuss the work of Kuzmin \emph{et al.}~\cite{kuzmin2017}, who also use traditional matching costs as well as a largely traditional pipeline for aggregation, using a learned deep network to control the parameters of this aggregation in different regions. This allows them to achieve a low run-time of 0.034 seconds (i.e., 29.4 frames per second) but with worse accuracy. Meanwhile, the DispNetC architecture introduced in \cite{sceneflow} uses 2D convolutions, like us, for spatial reasoning---applied on a feature map derived from computing cross-correlations between per-image feature maps at different disparity shifts. This also leads to reduced run-times (0.06 seconds in their case) but lower accuracy. Our method is able to achieve higher accuracy as well as lower run-times (0.021 seconds) than both these methods.

\begin{figure*}[!t]
  \centering
  \includegraphics[width=\textwidth]{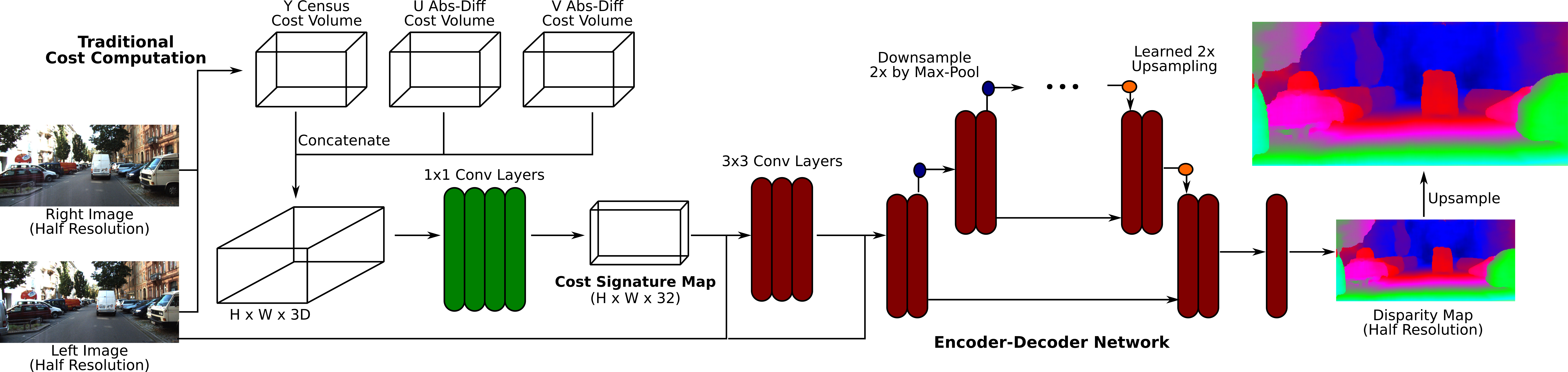}
  \caption{Proposed Stereo Estimation Pipeline. We propose a new computationally efficient architecture that avoids the use of learned matching costs and expensive 3D convolutions. We begin by constructing cost volumes at half-resolution by using different efficient traditional matching costs. We then generate a 2D cost-signature map using per-pixel layers to summarize the different costs at all disparities into a low-dimensional feature vector at each location. This is followed by spatial processing with an encoder-decoder architecture to produce an estimate of the final disparity map. Both the cost-signature and spatial processing layers are trained end-to-end.}
  \label{fig:arch}
\end{figure*}

\section{Proposed Method}
\label{sec:method}

We now describe our processing pipeline and network architecture for stereo estimation, summarized also in Fig.~\ref{fig:arch}. Our overall approach is based on avoiding expensive repeated floating point computations across different disparity candidates (i.e., 3D convolutions across a cost volume). To this end, we leverage multiple traditional efficient matching costs to build initial cost volumes, summarize this information in to a 2D feature map with a low-dimensional feature vector at each pixel location, and use 2D convolutions within an encoder-decoder architecture to yield the final disparity estimates. All our computations are also carried out at half resolution (as is common) for further efficiency, with a final interpolation step in the end to yield a disparity map at the original resolution.

\subsection{Initial Matching with Traditional Costs}

The first step in our pipeline considers the left and right images downsampled by a factor of 2x, and builds multiple cost volumes $\{C_i\}$, one for each kind of cost, of size $H\times W\times D$ where $H,W$ are the height and width of the image, and $D$ is the size of the set of disparity values (we consider integer disparities going from $0$ to $127$ pixels at the half resolution). Each element $C_i(x,y,d)$ of each volume measures the dis-similarity between pixels or regions at $(x,y)$ in the left image and $(x-d,y)$ in the right image (when $x-d < 0$, we fill in values from the first valid pixel for that $d$ in the same row $y$).

We construct three volumes after converting the images from RGB to YUV color space. The first volume is $C_1$ based on the distances between $5\times 5$ census transforms~\cite{census} of the Y (i.e., luminance) channels of the left and right images. This cost was preferred in traditional stereo algorithms due to its robustness to global intensity scaling, and its computational efficiency---it requires computing a census code once for each location in the left and right images based on local comparisons, following which all entries of the cost volume can be computed by computing the hamming distance between corresponding codes (which simply requires counting the number of 1s in the bit-wise XOR of the two codes). To exploit color information, the other two cost volumes $C_2$ and $C_3$ are populated with the absolute difference of the U and V values of the corresponding pixels in the left and right image. We normalize each volume separately so that their costs have zero mean and unit-variance (as determined across volumes created from images in a training set).

\subsection{Conversion to Per-Location Cost Signatures}

The next stage of our pipeline converts the set of different matching costs at all candidate disparities for each reference pixel $(x,y)$ into a succinct low-dimensional cost signature $S(x,y) \in \mathbb{R}^{32}$. We begin by concatenating the three cost volumes into one 3D tensor of size $H\times W\times (3D)$. This treated as a 2D feature map $A_0(x,y) \in \mathbb{R}^{3D}$, where each $A_0(x,y)$ contains a vector of all costs at all disparities:
\begin{equation}
  \label{eq:concat}
  A_0(x,y) = [C_1(x,y,0),\ldots C_1(x,y,D-1),C_2(x,y,0),\ldots],
\end{equation}
and learn a mapping from $A_0$ to the lower-dimensional $S_0$. This is different from other neural network-based approaches~\cite{3dconv,edgestereo,pyramid,pds,iresnet} that construct 4D tensors corresponding to 3D feature maps, to enable operations between neighboring disparities. We also do not combine the different costs with a weighted sum as in traditional stereo methods~\cite{SPS-St,chakrabarti2015low}, instead letting the cost-signature $S(x,y)$ be a more general functions of different costs at different disparities.

The dimensionality reduction from $3D$ (which is 384 in our setting of three costs and 128 candidate disparities) in $A_0$ to 32 in $S$ is carried out by four layers that operate independently on each pixel location, and reduce dimensions to 192,96,48, and finally 32 feature channels. We instantiate these as $1\times 1$ convolution layers, and use batch normalization~\cite{bnorm} and ReLU activations at the output of each layer.

\subsection{Spatial Processing of Cost Signatures}

The final stage of our pipeline uses a neural network with 2D convolution layers to estimate a disparity map from the cost-signature feature map $S$. We begin by concatenating the left input image (at half resolution) to the cost-signature map, and sending these through an initial set of three convolution layers---all with $3\times 3$ kernels, 32 channel outputs, and batch normalization and ReLU activations. We then take this output, concatenate it again with the left image, and feed it to a UNet~\cite{unet}-like encoder-decoder architecture.

This encoder-decoder network features five levels of downsampling in the encoder and upsampling in the decoder (each time by a factor of 2), with skip-connections (joined by concatenation) between corresponding scales of the encoder to the decoder. We use two $3\times 3$ convolution layers each in the decoder and encoder at each scale, and achieve downsampling by a $2\times 2$ max-pool operation with stride 2, and upsampling by a learned convolution layer. While the original UNet~\cite{unet} recommends doubling the number of feature channels at each scale, we choose to only increase the channels by 16 each time to reduce computation. We do not use batch-normalization in these layers.

The output of the final decoder layer is a 32-channel feature map at half-resolution. We use a single per-pixel layer to map this to a single channel disparity map, and then upsample this map to the full resolution. During training, we simply use nearest neighbor upsampling, while at test time we use a simple discontinuity-aware interpolation scheme: we produce both nearest-neighbor and bilinearly interpolated disparity maps, and at each pixel select the interpolated version when the difference between the two is less than 1 pixel, and the nearest-neighbor version otherwise.

\subsection{Training Loss}

Note that the first stage of our pipeline (\ref{sec:method}.A) is fixed and need not be learned. We train the cost-signature (\ref{sec:method}.B) and spatial processing (\ref{sec:method}.C) layers end-to-end based on a loss defined on the quality of the final full-resolution disparity map (i.e., after upsampling). We use a robust regression loss, between estimated and ground-truth disparities $\hat{d}$ and $d_{GT}$:
\begin{equation}
  \label{eq:loss}
  L(d_{GT},\hat{d}) = \max(\tau,d_{GT}-\hat{d})^{1/8}.
\end{equation}
The sub-linear exponent makes the loss robust to outliers (i.e., where the disparity error is too high), with clipping by $\tau$ used to ensure that gradients of the loss are stable (we set $\tau=1$ in our experiments). Note that the loss function is only computed over pixels with valid disparity values present in the ground truth.

\section{Experiments}
\label{sec:exp}

We implement our network architecture in Tensorflow~\cite{tensorflow}, using custom GPU operations for the initial cost volume computation in Sec.~\ref{sec:method}.A, and evaluate our method on the KITTI 2012~\cite{kitti} and 2015~\cite{kitti2} benchmarks. We report running times using an NVIDIA GTX 1080Ti GPU. For training, we adopt the standard practice of pre-training our network on the synthetic dataset of \cite{sceneflow}---specifically, on images from the ``FlyingThings3D'' and ``Driving'' sub-sets. We then fine-tune on images from the KITTI 2012 and 2015 training sets---although, we remove a subset of 20 images from the KITTI 2015 training set and use it for validation. Our reference implementation along with trained model weights will be made available at \url{https://projects.ayanc.org/fdscs/}.

\subsection{Training}

We train our network with the loss in \eqref{eq:loss} and weight decay of $10^{-5}$ using the Adam optimizer~\cite{adam}. We begin by training for 350k iterations on the synthetic dataset~\cite{sceneflow}, with a learning rate of $10^{-4}$ (after initially training for 5k iterations at a lower rate of $10^{-5}$ for stability). Although our network is designed to only produce disparities with respect to the co-ordinates of the left image, this dataset provides dense ground-truth disparity maps with respect to both the left and right images. To train also with the right image disparity map, we form an additional pair by swapping the left and right images and flipping both images and the disparity map horizontally. We train with a batch size of 4 original pairs, which yields an effective batch of 8 with the swapped pairs.

We then fine-tune on images from KITTI, and since this is a smaller dataset with sparse ground-truth data, we use scale augmentation: scaling the left, right, and ground truth images by a random scale factor in $(1.0,1.5)$, and dividing the ground truth disparities by the same value. We use a batch size of 4, and train for 150k iterations with a learning rate of $10^{-4}$, and additional 50k iterations at learning rates of $10^{-5}$ and $10^{-6}$ each.

\subsection{Ablation Study}

\begin{table}[!t]
  \centering
  \caption{Ablation Study on Validation Set}
  \label{tab:ablat}
  \renewcommand{\arraystretch}{1.25}
  \begin{tabular}{lccc}
    \hline
    Model & Avg Err. & \% Err. $> 3$px & Running Time\\\hline
    Full Model & 0.72 & 2.41 & 0.021s\\
    Only Census Cost & 0.75 & 2.60 & 0.016s\\
    Only 3-Level Enc-Dec & 0.90 & 3.65 & 0.020s\\\hline
  \end{tabular}
\end{table}

\begin{table*}[!t]
  \centering
  \caption{Results on the KITTI 2012~\cite{kitti} Benchmark}
  \label{tab:k2012}
  \renewcommand{\arraystretch}{1.25}
  \begin{tabular}{l|cc|cc|cc|cc|r}
    \hline
    & \multicolumn{2}{c|}{$>$ 2px} & \multicolumn{2}{c|}{$>$ 3px}& \multicolumn{2}{c|}{$>$ 4px} & \multicolumn{2}{c|}{$>$ 5px} & Run\\
    Method & Out-Noc & Out-All & Out-Noc & Out-All & Out-Noc & Out-All & Out-Noc & Out-All&Time\\\hline
    MC-CNN-acrt~\cite{zbontarlecun2} & 3.90& 5.45& 2.43& 3.63& 1.90& 2.85& 1.64& 2.39&67s\\
    GC-Net~\cite{3dconv}& 2.71& 3.46& 1.77& 2.30& 1.36& 1.77& 1.12& 1.46&0.9s\\
    Content-CNN~\cite{siamese}&4.98&6.51&3.07&4.29&2.39&3.36&2.03&2.82&0.7s\\
    SegStereo~\cite{segstereo}&2.66&3.19&1.68&2.03&1.25&1.52&1.00&1.21&0.6s\\
    PDSNet~\cite{pds}&3.82&4.65&1.92&2.53&1.38&1.85&1.12&1.51&0.5s\\
    PSMNet~\cite{pyramid}&2.44&3.01&1.49&1.89&1.12&1.42&0.90&1.15&0.41s\\
    EdgeStereo~\cite{edgestereo}& - & - & 1.73& 2.18& 1.30& 1.64& 1.04& 1.32& 0.27s\\
    iResNet-i2~\cite{iresnet}&2.69& 3.34& 1.71& 2.16& 1.30& 1.63& 1.06& 1.32&0.12s\\
    DispNetC~\cite{sceneflow}&7.38&8.11&4.11&4.65&2.77&3.20&2.05&2.39&0.06s\\
    DES-net~\cite{iresnet}&4.88&5.54& 2.66& 3.12& 1.78& 2.11& 1.33& 1.59&0.05s\\
    SPS-St~\cite{SPS-St} & 4.98 & 6.28 & 3.39 & 4.41 & 2.72 & 3.52 & 2.33 & 3.00 & (CPU) 2s\\
    \hline
    Proposed & 4.54 & 5.34 & 2.61 & 3.20 & 1.86 & 2.33 & 1.46 & 1.85&0.021s\\\hline
  \end{tabular}
\end{table*}

\begin{table*}[!t]
  \centering
  \caption{Results on the KITTI 2015~\cite{kitti2} Benchmark}
  \label{tab:k2015}
  \renewcommand{\arraystretch}{1.25}
  \begin{tabular}{l|ccc|ccc|r}
    \hline
    & \multicolumn{3}{c|}{All Pixels} & \multicolumn{3}{c|}{Non-Occluded Pixels}&Run\\
    Method & D1-bg & D1-fg & D1-all & D1-bg & D1-fg & D1-all & Time\\\hline
    MC-CNN-acrt~\cite{zbontarlecun2} & 2.89 & 8.88 & 3.89 & 2.48 & 7.64 & 3.33 & 67s\\
    GC-Net~\cite{3dconv} & 2.21 & 6.16 & 2.87 & 2.02 & 5.58 & 2.61 & 0.9s\\
    Content-CNN~\cite{siamese} & 3.73 & 8.58 & 4.54 & 3.32 & 7.44 & 4.00 & 0.7s\\
    SegStereo~\cite{segstereo} & 1.88 & 4.07 & 2.25 & 1.76 & 3.70 & 2.08 & 0.6s\\
    PDSNet~\cite{pds} & 2.29&4.05&2.58&2.09&3.68&2.36&0.5s\\
    PSMNet~\cite{pyramid} & 1.86 & 4.62 & 2.32 & 1.71 & 4.31 & 2.14 & 0.41s\\
    EdgeStereo~\cite{edgestereo} & 2.27 & 4.18 & 2.59 & 2.12 & 3.85 & 2.40 & 0.27s\\
    iResNet-i2~\cite{iresnet} & 2.25 & 3.40 & 2.44 & 2.07 & 2.76 & 2.19 & 0.12s\\
    DispNetC~\cite{sceneflow} & 4.32 & 4.41 & 4.34 & 4.11 & 3.72 & 4.05 & 0.06s\\
    DES-net~\cite{iresnet} & 3.13 & 3.87 & 3.25 & 2.94 & 3.21 & 2.98 & 0.05s\\
    DeepCostAggr~\cite{kuzmin2017} & 5.34 & 11.35 & 6.34 & 4.82 & 10.11 & 5.69 & 0.034s\\
    SPS-St~\cite{SPS-St} & 3.84 & 12.67 & 5.31 & 3.50 & 11.61 & 4.84 & (CPU) 2s\\
    \hline
    Proposed & 2.83 & 4.31 & 3.08 & 2.53 & 3.74 & 2.73 & 0.021s\\\hline
  \end{tabular}
\end{table*}

We begin by comparing our full approach on the validation set to different ablated versions in Table~\ref{tab:ablat}. Specifically, we consider a version of our model that only uses the census cost volume and leaves out the chromaticity difference-based costs, and a versions with a smaller networks for spatial processing with only three (instead of five) scales in the encoder-decoder. We report running times (for 1240x375 images) for these versions and  accuracy in terms of the average error (absolute difference between true and estimated disparity) as well as percentage of pixels where this error is greater than 3 pixels.

We see that both variations from our full model lead to higher errors and lower running times, but by different amounts. In particular, removing the color matching costs leads to a significant improvement in speed (from 48 to 62 frames per second) but only a modest drop in performance. In contrast, using a smaller spatial processing network leads to a barely measurable performance in speed but a significant drop in performance. This demonstrates that more accurate match information provides comparatively less value given the cost of larger number of computations across the disparity dimension. In contrast, more complex spatial processing with a larger receptive field makes a greater contribution to reliable estimation while being relatively cheap computationally since it involves only 2D processing.

\subsection{Results on KITTI Benchmark}

Finally, we report the official results as reported by the test server on the KITTI 2012 and 2015 benchmarks along with running time in Tables \ref{tab:k2012} and \ref{tab:k2015} respectively, and compare these to recent published works as well as a traditional (non neural network-based) method~\cite{SPS-St}. For the KITTI 2012 benchmark, errors are measured as percentage of pixels with disparity error above different thresholds, computed over non-occluded (Out-Noc) and all (Out-all) pixels. For KITTI 2015, all errors correspond to percentage of pixels with errors greater than 3 pixels for non-occluded and all pixels, reported separately for all pixels (D1-all) and those corresponding to background (D1-bg) and foreground (D1-fg) objects. We also include example estimated disparities and errors on select KITTI 2015 test images for our method and those from methods with relatively low running times, namely Deep Cost Aggregation~\cite{kuzmin2017} and DispNetC~\cite{sceneflow}, and those with high accuracy, SegStereo~\cite{segstereo} and PSMNet~\cite{pyramid}.

We find that our method has a clear advantage in accuracy over the traditional stereo method of \cite{SPS-St}, as well as over the only other method with real-time performance~\cite{kuzmin2017}. On KITTI 2015, it also performs better than the other methods, \cite{sceneflow} and the DES-net version of \cite{iresnet}, that take less than 0.1 seconds per stereo pair, although it performs slightly worse than DES-net on KITTI 2012. At the same time, it's performance is competitive to state-of-the-art methods~\cite{segstereo,pyramid}: on the D1-all metric on non-occluded pixels in KITTI 2015, it is worse by only 0.65\% and 0.59\% to SegStereo and PSMNet, respectively, while being between 20-30 times faster. Therefore, our method provides a new trade-off in accuracy vs speed in stereo estimation, and permits practical real-time estimation with competitive accuracy.

\begin{figure*}[!t]
  \centering
  \begin{tabular}{cccccc}
    & Input Stereo Pair & & Proposed Method (48 FPS) & DeepCostAggr~\cite{kuzmin2017} (29 FPS)\\
    \rotatebox{90}{\small ~~~~Left}\rotatebox{90}{\small ~~~Image}\hspace{-1em} & \includegraphics[height=4.5em]{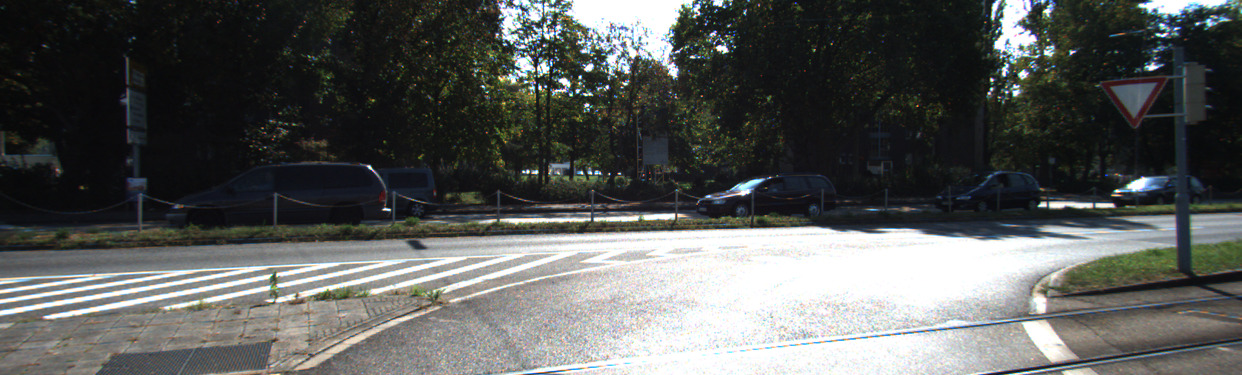} & \rotatebox{90}{\small ~~Disparity}\hspace{-1em} & \includegraphics[height=4.5em]{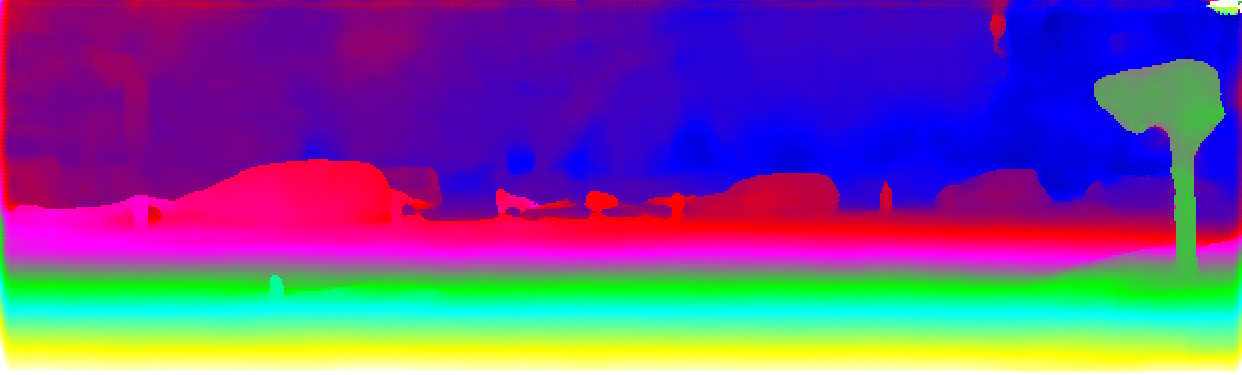} & \includegraphics[height=4.5em]{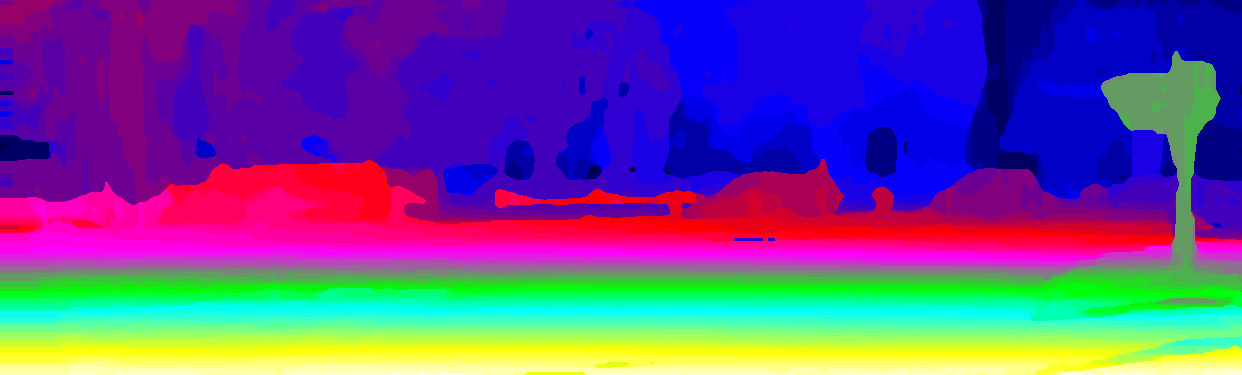}\\
    \rotatebox{90}{\small ~~~~Right}\rotatebox{90}{\small ~~~Image}\hspace{-1em} & \includegraphics[height=4.5em]{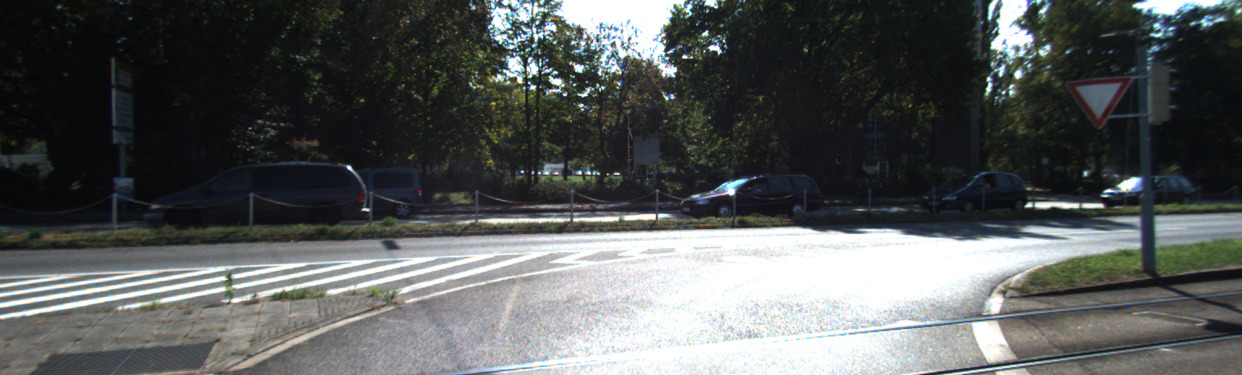} & \rotatebox{90}{\small ~~~~Errors}\hspace{-1em} & \includegraphics[height=4.5em]{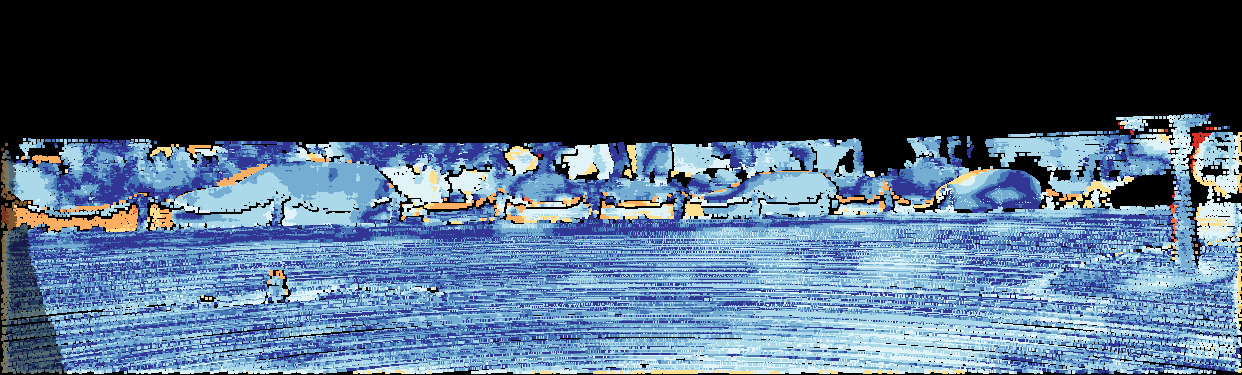} & \includegraphics[height=4.5em]{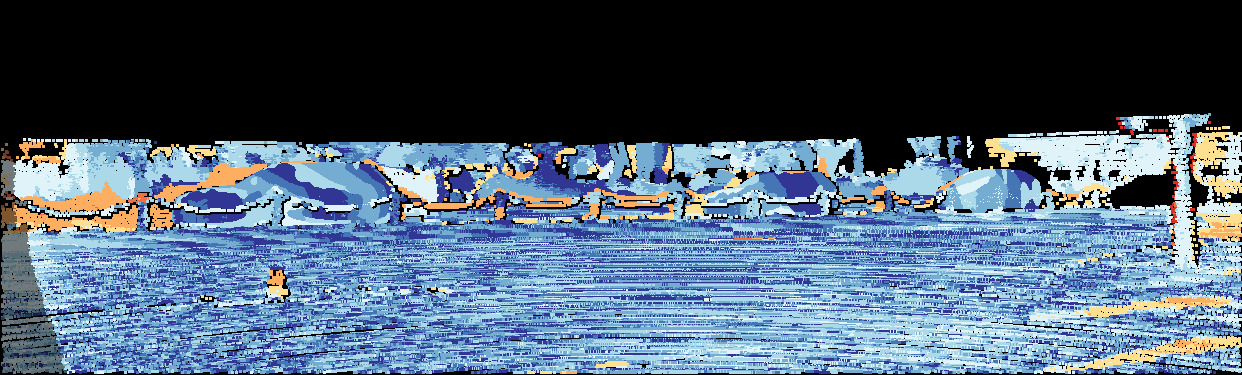}\\
    & DispNetC~\cite{sceneflow} (17 FPS) &  & PSMNet~\cite{pyramid} (2.4 FPS) & SegStereo~\cite{segstereo} (1.7 FPS)\\
    \rotatebox{90}{\small ~~Disparity}\hspace{-1em} & \includegraphics[height=4.5em]{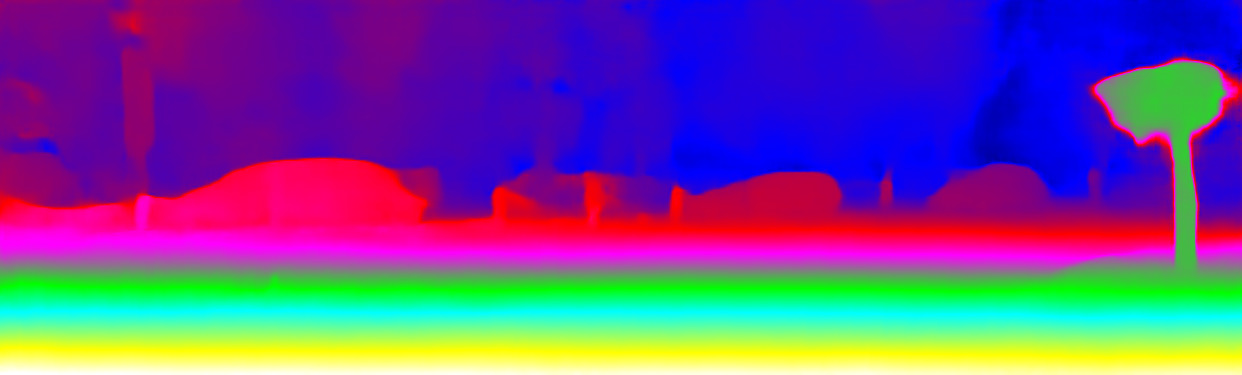} & & \includegraphics[height=4.5em]{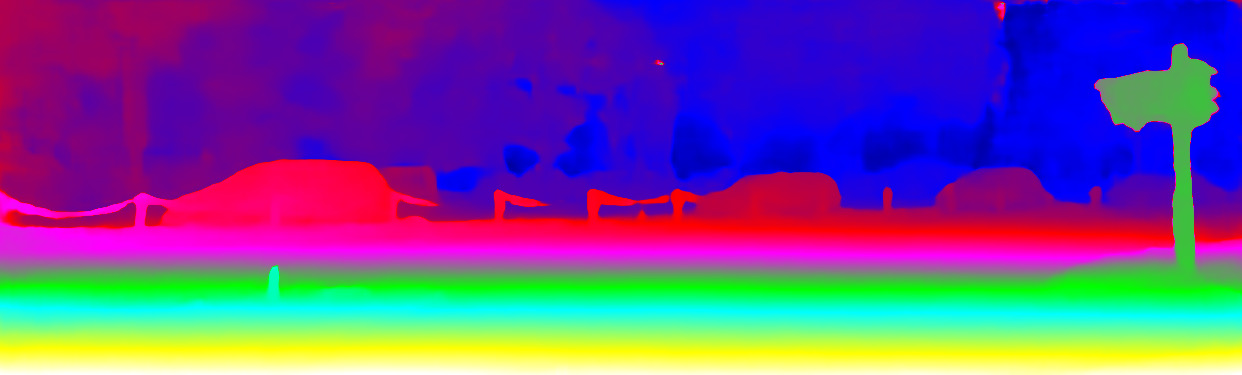} & \includegraphics[height=4.5em]{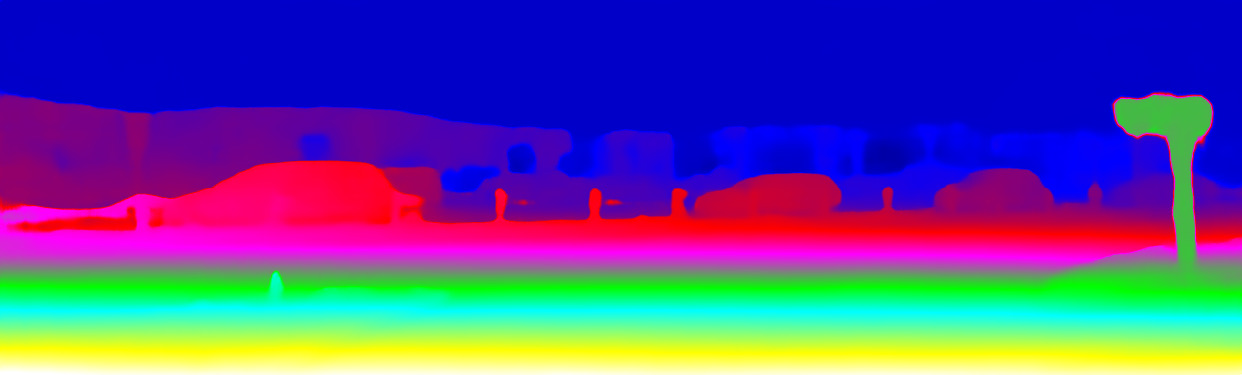}\\
    \rotatebox{90}{\small ~~~~Errors}\hspace{-1em} & \includegraphics[height=4.5em]{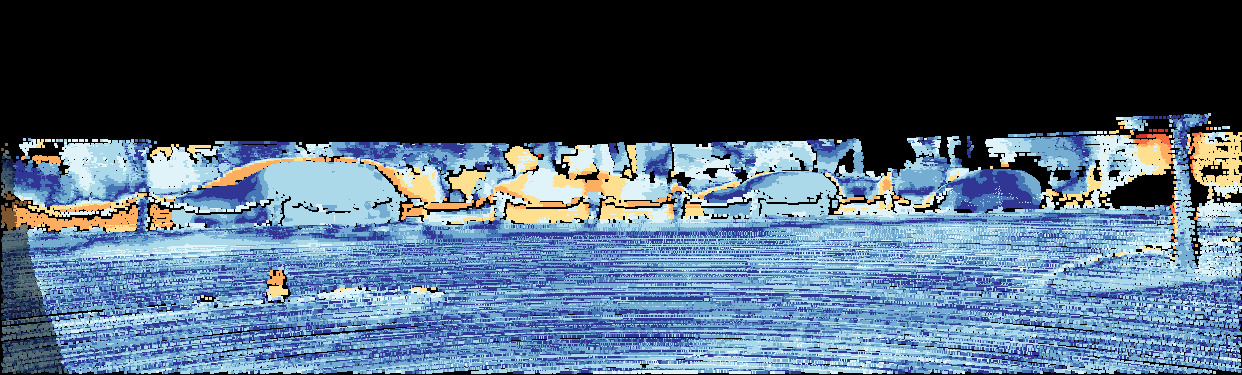} & & \includegraphics[height=4.5em]{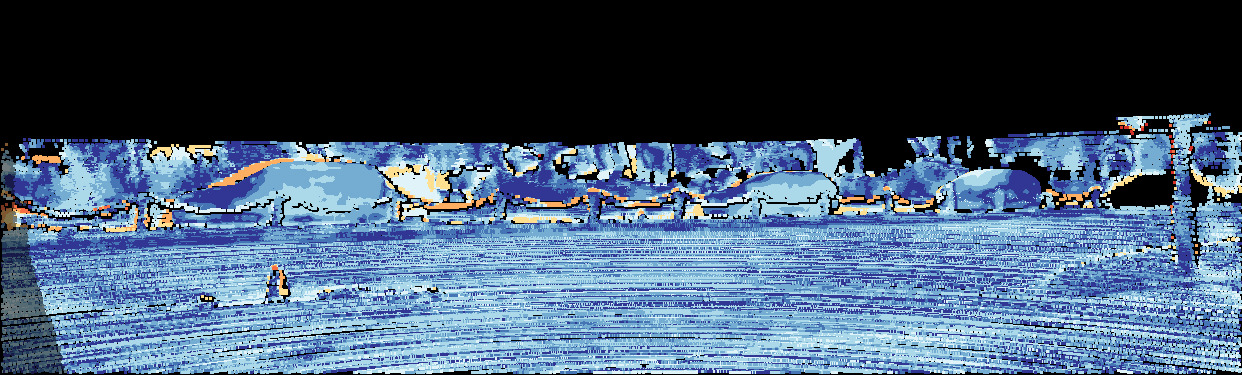} & \includegraphics[height=4.5em]{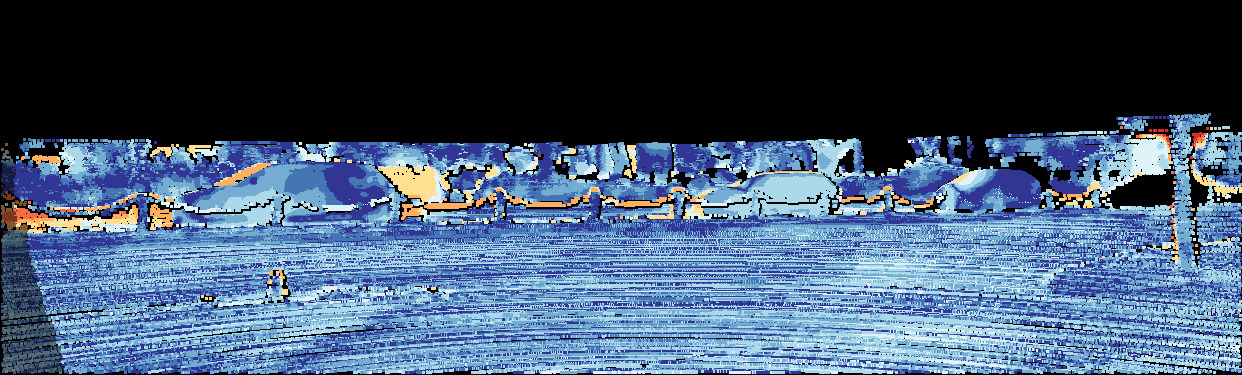}\\

    \vspace{-0.5em}\\\hline\vspace{-0.5em}\\

    & Input Stereo Pair & & Proposed Method (48 FPS) & DeepCostAggr~\cite{kuzmin2017} (29 FPS)\\
    \rotatebox{90}{\small ~~~~Left}\rotatebox{90}{\small ~~~Image}\hspace{-1em} & \includegraphics[height=4.5em]{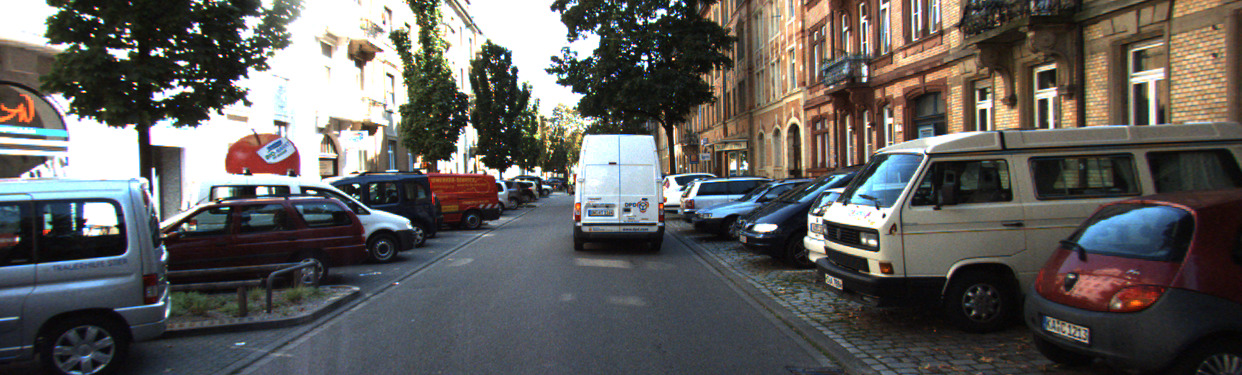} & \rotatebox{90}{\small ~~Disparity}\hspace{-1em} & \includegraphics[height=4.5em]{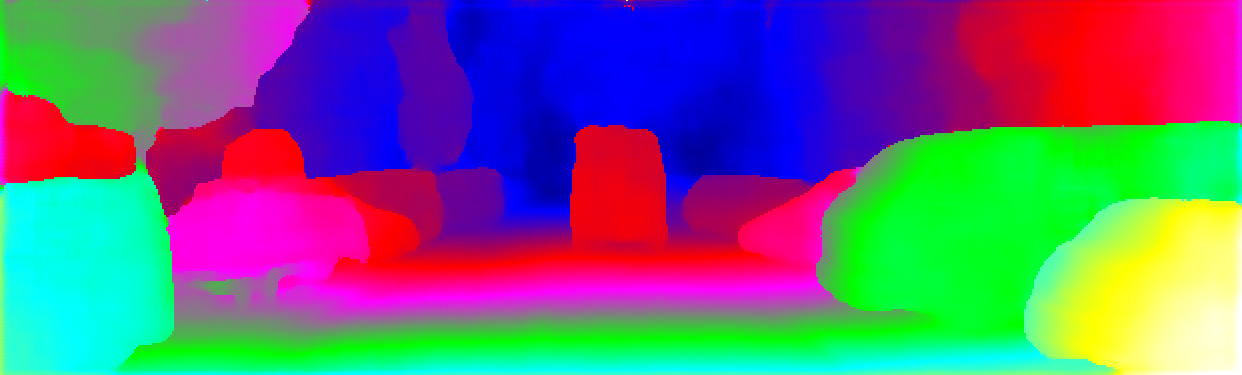} & \includegraphics[height=4.5em]{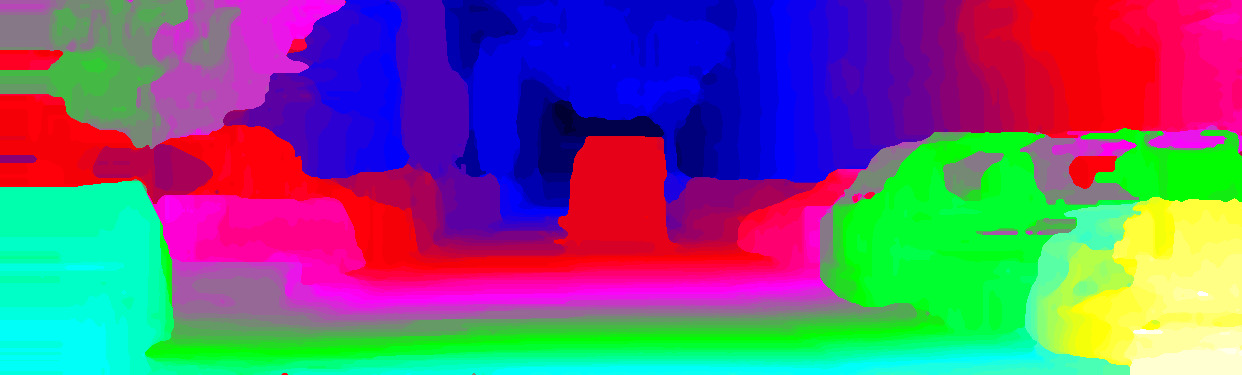}\\
    \rotatebox{90}{\small ~~~~Right}\rotatebox{90}{\small ~~~Image}\hspace{-1em} & \includegraphics[height=4.5em]{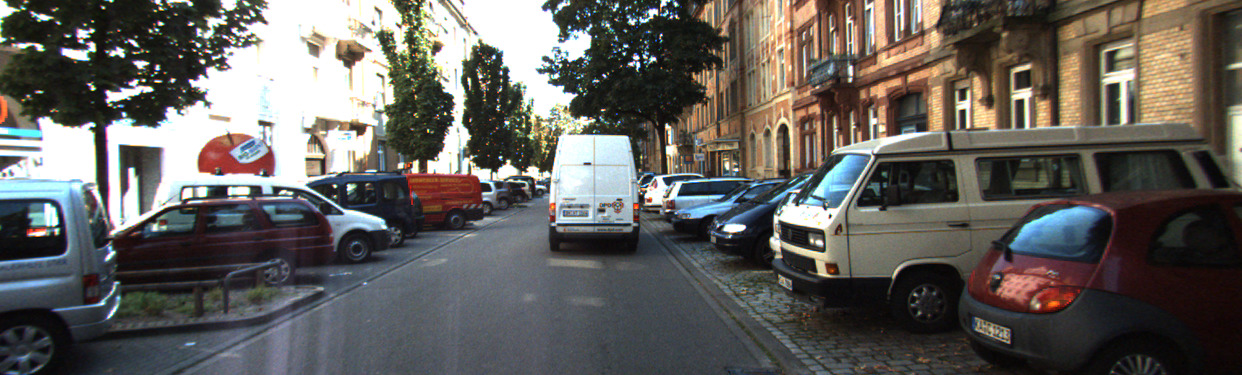} & \rotatebox{90}{\small ~~~~Errors}\hspace{-1em} & \includegraphics[height=4.5em]{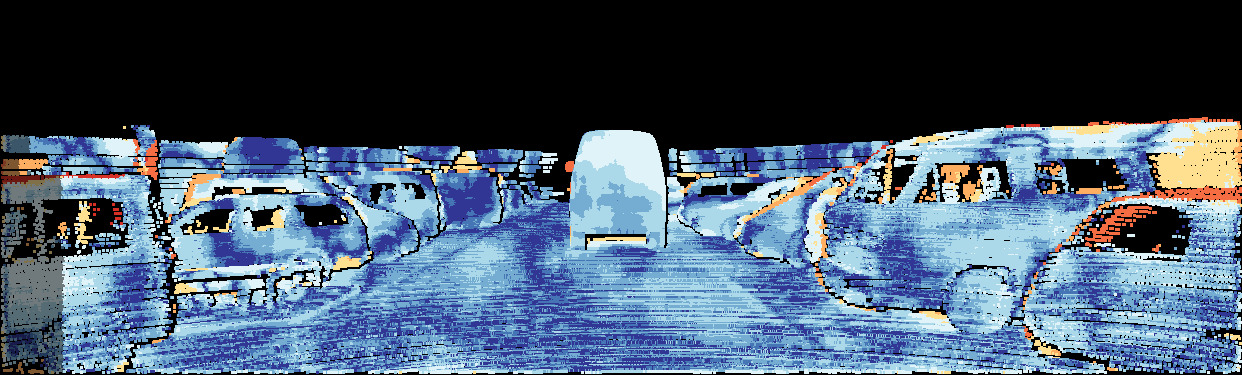} & \includegraphics[height=4.5em]{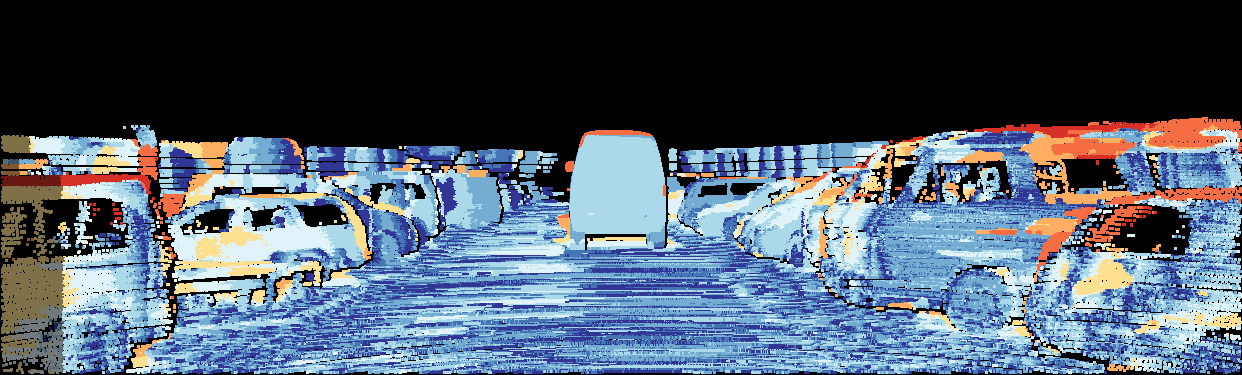}\\
    & DispNetC~\cite{sceneflow} (17 FPS) &  & PSMNet~\cite{pyramid} (2.4 FPS) & SegStereo~\cite{segstereo} (1.7 FPS)\\
    \rotatebox{90}{\small ~~Disparity}\hspace{-1em} & \includegraphics[height=4.5em]{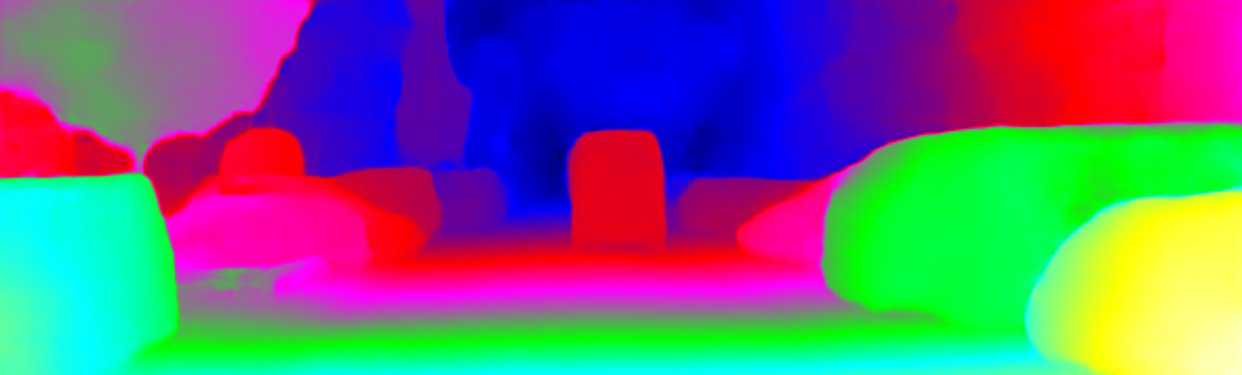} & & \includegraphics[height=4.5em]{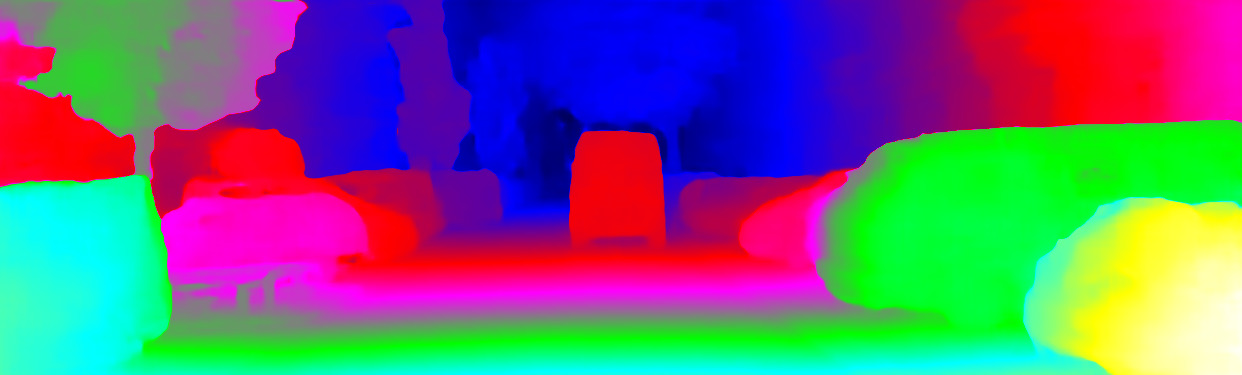} & \includegraphics[height=4.5em]{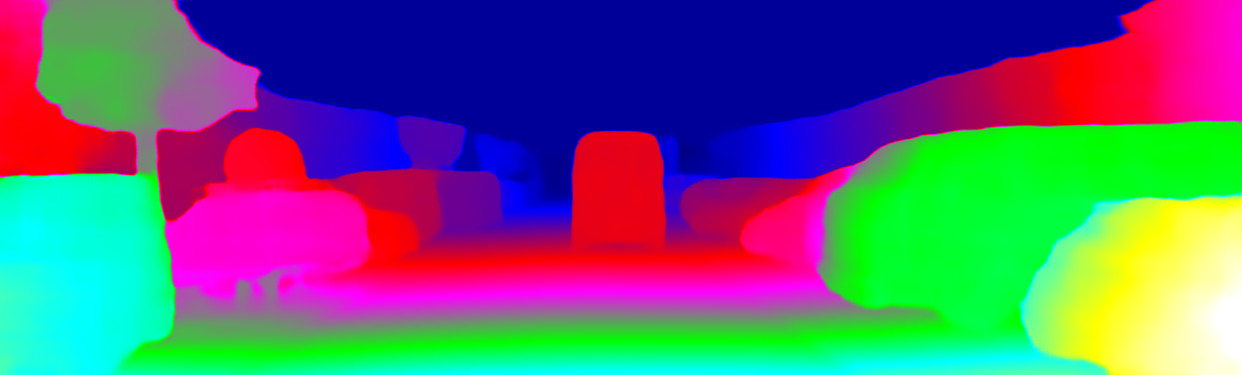}\\
    \rotatebox{90}{\small ~~~~Errors}\hspace{-1em} & \includegraphics[height=4.5em]{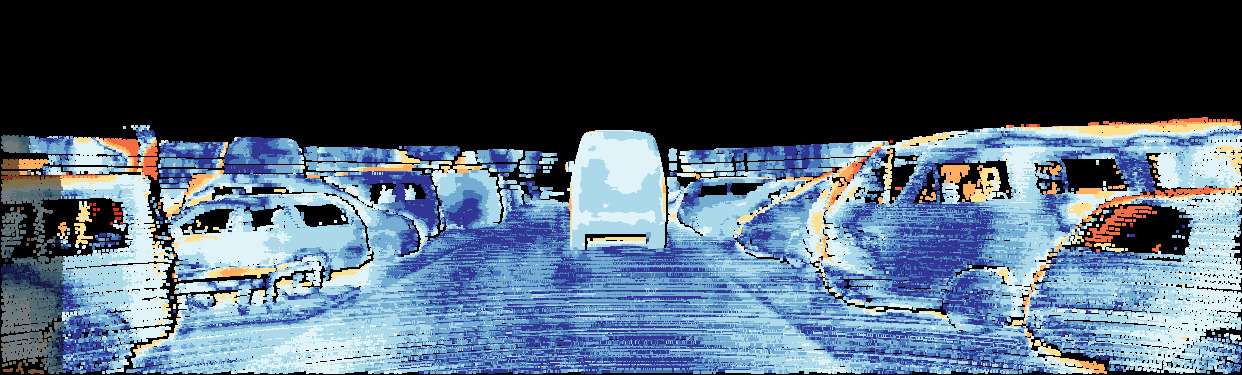} & & \includegraphics[height=4.5em]{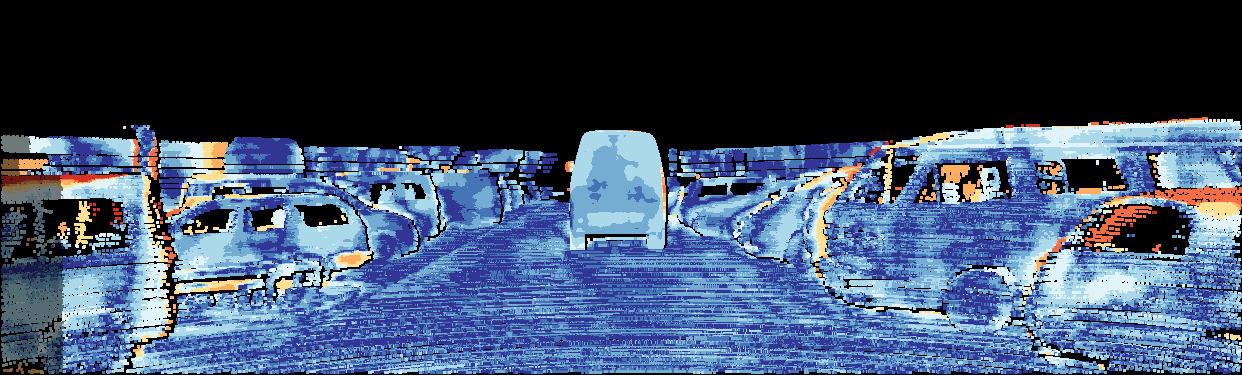} & \includegraphics[height=4.5em]{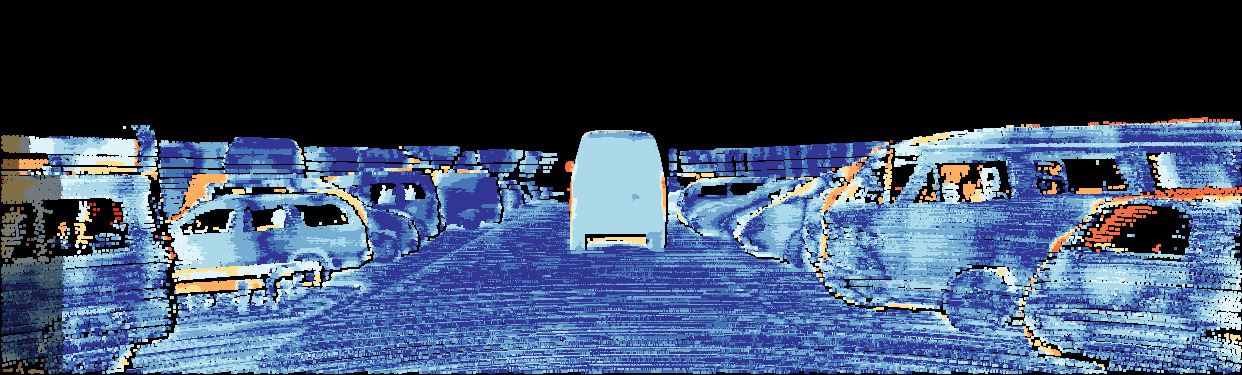}\\

  \end{tabular}
  \caption{Example Results on KITTI 2015~\cite{kitti2} test images. We show disparity maps, estimated by our model as well other methods, and corresponding errors on example stereo pairs from the KITTI 2015 test set. Disparity and error maps are shown using the standard color scheme of the benchmark.}
  \label{fig:results}
\end{figure*}

\section{Conclusion}
\label{sec:conc}

We introduced a new stereo estimation method that is able to generate accurate dense depth estimates from stereo image pairs at faster than real time speeds---48 frames per second---on a modern GPU. Our method achieves this by using traditional matching costs instead of their more expensive learned counterparts, and focusing its computations on spatial processing with 2D convolutions, in contrast to recent neural network-based methods that seek to explicitly mimic the cost-volume computations of traditional stereo pipelines. Given its accuracy and speed, our method is feasible to deploy on actual robots and autonomous vehicles, possibly as an alternative for depth perception to more expensive LIDARs. While our work focused on the binocular stereo case in this paper, in future work we propose extending our approach to multi-view stereo---we believe the higher computational efficiency will make it possible to reason about correspondences across multiple cameras in real time, while bringing gains in depth estimation accuracy.

\section*{Acknowledgments}
This work was supported by the NSF under award no. IIS-1820693, including with an REU supplement for KY's participation in the project at Washington University.
\nocite{kitti}


\end{document}